\documentclass{llncs}
\usepackage{array,multirow,graphicx}
\usepackage{balance}
\usepackage[english]{babel}
\usepackage{lipsum}
\usepackage{amssymb}
\usepackage{enumitem}
\usepackage{xfrac}
\usepackage{cite}
\usepackage{amsmath,amssymb,amsfonts}
\usepackage{algorithmic}
\usepackage{graphicx}
\usepackage{textcomp}
\def\BibTeX{{\rm B\kern-.05em{\sc i\kern-.025em b}\kern-.08em
    T\kern-.1667em\lower.7ex\hbox{E}\kern-.125emX}}
\usepackage[utf8]{inputenc}
\usepackage{csquotes}
\usepackage{url}
\usepackage{stfloats}
\usepackage{amsmath}
\usepackage{cases}
\usepackage{array}
 \usepackage{pifont}
\usepackage[para]{footmisc}
\newcolumntype{P}[1]{>{\centering\arraybackslash}p{#1}}

\begin{document}

\title{The Regression Tsetlin Machine: A Tsetlin Machine for Continuous Output~Problems \vspace{-3mm}}

\author{K. Darshana Abeyrathna \and
Ole-Christoffer Granmo \and
Lei Jiao \and Morten~Goodwin}
\institute{Centre for Artificial Intelligence Research, University of Agder, Grimstad, Norway
\email{\{darshana.abeyrathna, ole.granmo, lei.jiao, morten.goodwin\}}@uia.no}
\maketitle 

\begin{abstract}
The recently introduced Tsetlin Machine (TM) has provided competitive pattern classification accuracy in several benchmarks, composing patterns with easy-to-interpret conjunctive clauses in propositional logic. In this paper, we go beyond pattern classification by introducing a new type of TMs, namely, the \emph{Regression Tsetlin Machine} (RTM). In all brevity, we modify the inner inference mechanism of the TM so that input patterns are transformed into a single continuous output, rather than to distinct categories. We achieve this by: (1) using the conjunctive clauses of the TM to capture arbitrarily complex patterns; (2) mapping these patterns to a continuous output through a novel voting and normalization mechanism; and (3) employing a feedback scheme that updates the TM clauses to minimize the regression error. The feedback scheme uses a new activation probability function that stabilizes the updating of clauses, while the overall system converges towards an accurate input-output mapping. The performance of the RTM is evaluated using six different artificial datasets with and without noise, in comparison with the Classic Tsetlin Machine (CTM) and the Multiclass Tsetlin Machine (MTM). Our empirical results indicate that the RTM obtains the best training and testing results for both noisy and noise-free datasets, with a smaller number of clauses. This, in turn, translates to higher regression accuracy, using significantly less computational resources.
\vspace{-3mm}
\end{abstract}


\section{Introduction}
Computational simplicity, ease of interpretation, along with competitive pattern recognition accuracy, make the recently introduced Tsetlin Machine (TM) \cite{Ole1} a promising new paradigm for machine learning. Indeed, the TM has outperformed well-known machine learning algorithms such as Logistic Regression, Neural Networks, and Support Vector Machine (SVM) in several benchmarks, including Iris Data Classification, Handwritten Digits Classification (MNIST), Predicting Optimum Moves in the Axis and Allies Board Game, and Classification of Noisy XOR Data with Non-Informative Features \cite{Ole1}.

{\bf Tsetlin Automata and the Tsetlin Machine.} The core of the TM is built on Tsetlin Automata (TAs), developed by M. L. Tsetlin in the early 1960s \cite{tsetlin10}. This powerful, yet simple, leaning mechanism has been used to solve a number of machine learning and stochastic optimization problems, such as resource allocation \cite{granmo16}, stochastic searching on the line \cite{oommen19}, distributed coordination \cite{tung17}, graph coloring \cite{bouhmala18}, and forecasting disease outbreaks \cite{darshana5}. In the TM, TAs represent literals -- input features and their negations. The literals, in turn, form conjunctive clauses in propositional logic, as decided by the TAs. The final TM output is a disjunction of all the specified clauses. In this manner, the pattern composition and learning procedure of the TM is fully transparent and understandable, facilitating human interpretation. In addition, the TM has an inherent computational advantage. That is, the inputs and outputs of the TM can naturally be represented as bits, and recognition and learning is performed by manipulating those bits. The operation of the TM thus demands relatively small computational resources, and supports hardware-near and parallel computation e.g. on GPUs.

Lately, the TM has provided state-of-the-art performance in several real-life applications. Berge et al. have for instance successfully used the TM for medical text categorization \cite{Geir7}. They used the TM to provide interpretable pattern recognition for the analysis of electronic health records. The authors demonstrated that the TM can outperform established machine learning algorithms such as k-nearest neighbors (kNN), SVM, Random Forest, Decision Trees, Multilayer Perceptron (MLP), Long Short-Term Memory (LSTM) Neural Networks, and Convolutional Neural Networks (CNNs), in terms of precision, recall, and F-measure. Furthermore, Darshana et al. have shown that the TM can outperform MLPs, Decision Trees, and SVMs in dengue fever outbreak prediction. The latter result was achieved by making the TM capable of expressing thresholds and intervals that capture patterns formed by continuous features. By carefully selecting thresholds and intervals, the TM avoided losing information due to binarization \cite{darshana13}. 

{\bf Research Question and Paper Contributions.} The TM has been designed for classification, not for producing continuous output. How to best produce continuous output is unclear, with the existing binarization schemes being incapable of fully leveraging the natural ranking of numbers. In this paper, we introduce the Regression Tsetlin Machine (RTM) to overcome above limitation of the TM. The RTM is a novel variant of the Classic Tsetlin Machine (CTM), specifically addressing the unique properties of regression. The novel modifications that we introduce are subtle, but crucial. First of all, the clause polarities the CTM uses to discriminate patterns, using positive and negative examples, are eliminated. Instead, the objective of the RTM is to use the clauses to map the sum of the clause outputs into one single continuous output. The discrepancy between predicted and target output is minimized with a new feedback scheme tailored for regression, including a modified stochastic activation probability function.

{\bf Paper Organization.} The remainder of the paper is organized as follows. In Section~2, we present the main contribution of this paper, which is the RTM, and how we build it upon the CTM. We then investigate the behavior of the RTM using six different artificial datasets in Section 3. We  demonstrate empirically that the RTM is superior both to the CTM as well as its multiclass version when it comes to predicting continuous output. We conclude our work in Section 4.

\section{The Regression Tsetlin Machine (RTM)}

The RTM is a novel variant of the CTM. To highlight the unique properties of the RTM, we start this section with first reviewing the TM in more detail, and then discuss how it can be modified to support continuous output. 

\subsection{The Classic Tsetlin Machine (CTM)}\label{sec2a}

At the heart of the TM, we find multiple teams of TAs that build conjunctive clauses in propositional logic. The purpose is to capture hidden patterns in the data.

{\bf Learning with TAs.} Each Tsetlin Automaton (TA) learns the optimal action in an environment by sequentially performing the actions that the environment offers. To identify the optimal actions, the TAs adjust their states based on the feedback they receive from the environment, which can be penalties or rewards. Asymptotically, a TA identifies the action that provides the highest probability of reward \cite{misra3,tuan4}.  These simple learning devices are capable of online learning, have a simple structure, and require modest computational power. Yet, they are able to learn accurately with relatively few interactions with the environment \cite{narendra2,narendra9}. 

\begin{table*}[b]
\vspace{-3mm}
\caption{The steps used to form a clause based on the input features and the actions of the TAs.} \label{tab1}
\vspace{-5mm}
\begin{center}
\begin{tabular}{c}
\includegraphics[width=12.15cm ]{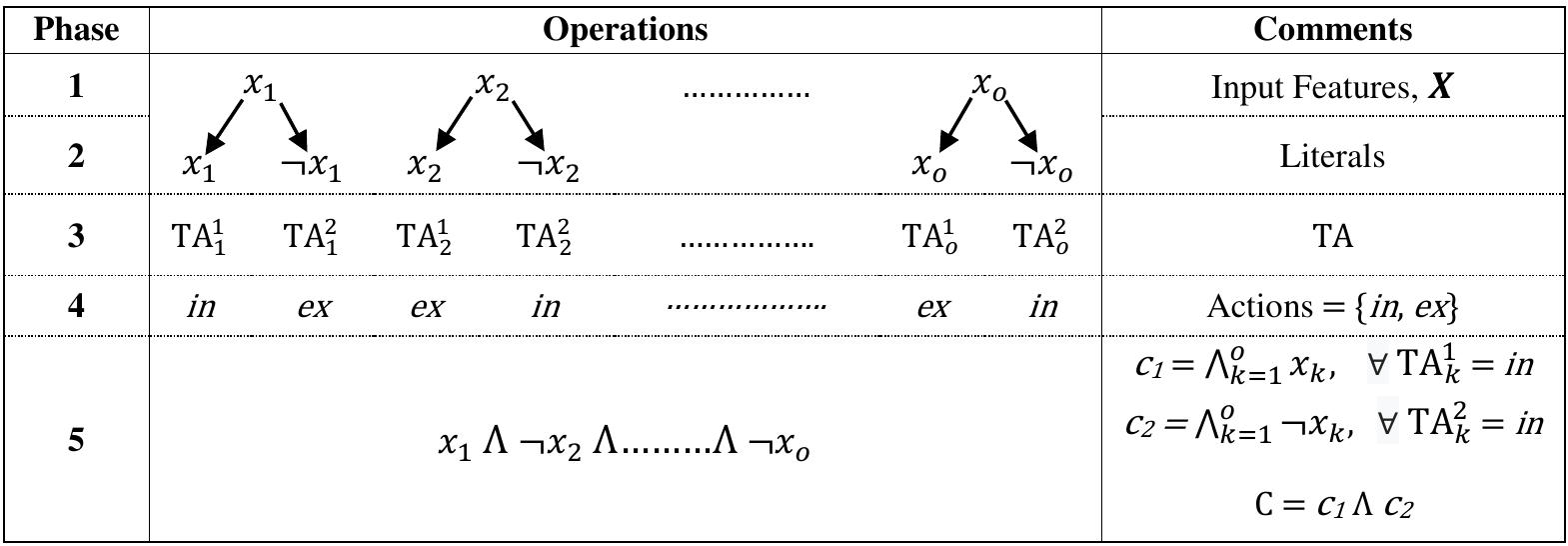}
\end{tabular}
\label{tab1}
\vspace{-7mm}
\end{center}
\end{table*}

{\bf Clause Formation and the TA Team.} The TM bases its operations on the simplest form of TAs, namely, the two action one, with finite memory depth. As illustrated in Table \ref{tab1}, a team of TAs cooperates to form a clause. The table depicts the steps leading to a clause being formed. Consider an input feature vector $\textbf{\textit{X}} = [x_1, x_2,\ldots, x_o]$. Each TA represents either an input feature $x_k$ or its negation $\lnot x_k$ (jointly referred to as literals). Further, each TA in the team decides whether to include or exclude its assigned literal in the clause that the team is forming. Accordingly, when there are $o$ input features, $2 \times o$ TAs are needed to form the clause. The two actions available to each TA are \{\textit{in}, \textit{ex}\}. Here, \emph{in} refers to including the literal controlled by the TA and \emph{ex} refers to excluding it. As seen in the final step in the table, the included literals form a conjunctive clause, while the excluded ones are ignored.

{\bf Clauses and Voting.}
The number of clauses, \textit{m}, needed for a particular problem depends on the complexity of the dataset. It should at least be sufficient to cover the full range of sub-patterns associated with each output \{0, 1\}. However, with hidden and unknown sub-patterns, a grid search is required to find the best \textit{m}.

The $m$ clauses are assigned either a positive or negative polarity, and they vote separately to decide the final output of the TM. Clauses with odd index are assigned positive polarity ($C^+$) and they vote for the final output 1. Clauses with even index are assigned negative polarity ($C^-$)  and they vote for the final output 0. For both categories, a vote is submitted when the clause recognizes a sub-pattern. If the clause is unable to find a sub-pattern, it declines to vote. Finally, the output, $y$, is decided based on the number of votes gained by each category \{0, 1\} as given in the Eq. (\ref{eq1}):
\vspace{-1mm}

\begin{equation}\label{eq1}
    y = 
\begin{cases}
    1          ,& \;\;\;\; \text{if } \;\; $$\sum_{j=1,3,m-1}$$ \;\; C_j^+ \; > \;\; $$\sum_{j=2,4,m} $$ \;\; C_j^-\\
    \\
    0,          & \;\;\;\; \text{if } \;\; $$\sum_{j=1,3,m-1}$$ \;\; C_j^+ \; < \;\; $$\sum_{j=2,4,m} $$ \;\; C_j^- \;\; \;\;.
\end{cases}
\vspace{1mm}
\end{equation}

{\bf Learning Procedure.} Learning in the TM is based on reinforcement learning. The reward, penalty, and inaction probabilities that guide the TAs in all of the clauses depend on several factors, namely, the actual output, the clause output, the literal value, and the current state of the TA. The basic idea is to alter the number of votes belong to each output category when the output is a false negative or a false positive. In the TM, this is done by two types of feedback – Type I and Type II. Type I feedback eliminates false negative output and reinforces true positive output, while Type II feedback eliminates false positive output. Both of these kinds of feedback are summarized in Table~\ref{tab2}.

\begin{table}[b]
\vspace{-4mm}
\centering
\newcolumntype{P}[1]{>{\centering\arraybackslash}p{#1}}
\caption{Type I and Type II feedback designed to eliminate false negative and false positive output.}\label{tab2}
\vspace{-2mm}
\begin{tabular}{|P{8mm}|c|c|c|c|c|c|P{6mm}|P{6mm}|P{6mm}|P{6mm}|}
\hline
\multicolumn{3}{|c|}{Feedback Type} & \multicolumn{4}{c|}{I} & \multicolumn{4}{c|}{II}  \\ \hline
\multicolumn{3}{|c|}{Clause Output} & \multicolumn{2}{c|}{1} & \multicolumn{2}{c|}{0} & \multicolumn{2}{c|}{1} & \multicolumn{2}{c|}{0} \\ \hline
\multicolumn{3}{|c|}{Literal Value} & 1 & 0 & 1 & 0 & 1 & 0 & 1 & 0 \\ \hline
\multirow{6}{*}{\rotatebox[origin=c]{90}{Current State}} & \multirow{3}{*}{Include} & Reward Probability & (s-1)/s & NA & 0 & 0 & 0 & NA & 0 & 0 \\
                        &                     & Inaction Probability & 1/s & NA & (s-1)/s & (s-1)/s & 1 & NA & 1 & 1\\
                        &                     & Penalty Probability & 0 & NA & 1/s & 1/s & 0 & NA & 0 & 0 \\ \cline{2-11}
                        & \multirow{3}{*}{Exclude} & Reward Probability & 0 & 1/s & 1/s & 1/s & 0 & 0 & 0 & 0\\
                        &                     & Inaction Probability & 1/s & (s-1)/s & (s-1)/s & (s-1)/s & 1 & 0 & 1 & 1\\
                        &                     & Penalty Probability & (s-1)/s & 0 & 0 & 0 & 0 & 1 & 0 & 0       \\ 
\hline
\end{tabular}
\vspace{-2mm}
\dag \hspace{1mm} s is the precision and controls the granularity of the sub-patterns captured \cite{Ole1}
\end{table}

Type I feedback is given to clauses with positive polarity when the actual output, $\hat{y}$, is 1 and clauses with negative polarity when the actual output, $\hat{y}$, is 0. The probability of activation of Type I feedback is {[$T - \max (-T, \min (T, $$\sum_{j=1}^m$$ C_j ))]/2T$}. Type II feedback is given to clauses with positive polarity when the actual output, $\hat{y}$, is 0 and clauses with negative polarity when the actual output, $\hat{y}$, is 1. The probability of activation of Type II feedback is {[$T + \max (-T, \min (T, $$\sum_{j=1}^m$$ C_j ))]/2T$}. TAs remain unchanged if the vote difference, {$\sum_{j=1}^m C_j$}, is higher than or equal to \textit{T} when $\hat{y}$ = 1 and lower than or equal to -\textit{T} when $\hat{y}$ = 0, according to the activation probabilities of each type of feedback.

In all brevity, when the target output for a training instance $\hat{X}$ is $\hat{y} = 1$, the votes from the clauses with negative polarity must not outnumber the votes from the clauses with positive polarity (in order to correctly classify the instance). Therefore, clauses with positive polarity receive Type I feedback (the activation probability increases with the number of voting clauses with negative polarity) since this reinforces clauses which output 1. Similarly, clauses with negative polarity receive Type II feedback (the activation probability increases with the number of voting clauses with positive polarity) since this suppresses voting activity by making clauses of negative polarity evaluate to 0. The procedure is similar when the target output is $\hat{y} = 0$. The TM then needs to make sure that more clauses with negative polarity provide votes compared to those with positive polarity. Eventually, the above feedback reduces the number of false positives and false negatives to make the TM learn the propositional formulae that provide high accuracy output. 

\subsection{The Multiclass Tsetlin Machine (MTM)}\label{sec2b}

For the CTM, the final summation operator aggregates all of the clause outputs into one of the two available outputs: $0$ or $1$. However, for categorization tasks with more classes than two, another design is needed. In the Multiclass Tsetlin Machine (MTM), clauses are partitioned equally among the classes. The clauses of each individual class then act separately, similarly to a single TM. However, the votes output for each class then form the basis for classification. That is, an argmax operator arbitrates the final class, based on the votes collected for each class. When there are $n$ classes, the output $y$ can thus be expressed as:

\begin{equation}
y = \mathrm{argmax}_{i=1,\ldots,n} \Bigg\{\Bigg(\sum_{j=1,3, \ldots (\frac{m}{n})-1}  C_j^i  \; - \sum_{j=2,4,\ldots (\frac{m}{n})} C_j^i\Bigg)\Bigg\}.
\vspace{4mm}
\end{equation}

The training procedure is similar to the CTM training procedure. However, in the MTM, the clauses of the class being the target of the current training sample are treated as if $\hat{y} = 1$, while the clauses of a randomly selected class from the remaining classes is treated as if $\hat{y} = 0$. In each class, clauses with positive polarity vote to say that the output belongs to the considered class. Similarly, the clauses with negative polarity vote to indicate that the output does not belong to the considered class.

\subsection{The Regression Tsetlin Machine (RTM)}\label{sec2c}

When the output is continuous, neither the CTM or the MTM above are ideal. However, we will now show that the CTM can be modified to produce continuous output by means of three pertinent modifications.

In CTM and MTM, the polarity of clauses is used to classify data into different classes. We now remove the polarity of clauses, since we intend to use the clauses as additive building blocks that can be used to calculate continuous output. That is, we intend to map the total vote count into a single continuous output. As a result, the complexity of the RTM is actually reduced.

With merely one type of clauses, the summation operator outputs a value between 0 and \textit{T}, which is simply the number of clauses that evaluates to 1. This value is then normalized to produce the regression output. Thus, through this simple modification, the TM can now produce continuous output, with precision that increases with higher \textit{T}.

Let $\hat{y}_{\mathrm{max}}$ denote the maximum output value $\hat{y}$ among the $N$ training samples $\textbf{\textit{Y}} = [\hat{y}_1, \hat{y}_2, \hat{y}_3, \ldots, \hat{y}_N]$. Then the sum of the votes from the clauses $\sum_{j=1}^m C_j$ of the TM is normalized to achieve the regression output by dividing by $T$ and multiplying with $\hat{y}_{\mathrm{max}}$. So, for the $o^{th}$ training sample, $(\hat{X}_o,\hat{y}_o)$, the TM output, $y_o$, is calculated from the input $\hat{X}_o$ as follows:
\begin{equation}\label{eq3}
\vspace{1mm}
y_o = \frac{\sum_{j=1}^m C_j(\hat{X}_o) \;\; \times \hat{y}_{\mathrm{max}}\;\; }{T} \;. 
\vspace{1mm}
\end{equation}

Feedback, then, is based on comparing the output, $y_o$ of the TM with the target output $\hat{y}_o$. The target value $\hat{y}_o$ can be higher or lower than the output value $y_o$. This is our basis for our new feedback scheme. That is, similarly to other machine learning methods, certain internal operations are needed to minimize the error between the predicted output, $y_o$, and target output, $\hat{y}_o$. In the RTM, this is quite simply achieved by providing Type I and Type II feedbacks according to the following criteria:

\begin{equation}\label{eq4}
    Feedback = 
\begin{cases}
    \text{Type I},   \;\;\;\;\;\;\;\;   & \text{if \;\; } y_o  <  \hat{y}_o \; ,\\
    \\
    \text{Type II},      \;\;\;\;\;\; \;\;   & \text{if \;\; } y_o > \hat{y}_o \;.
\end{cases}
\vspace{1mm}
\end{equation}

As with the CTM, the idea here is to increase the number of clauses that output 1 when the predicted output is less than the target output ($y_o < \hat{y}_o$). To achieve this, we then provide Type I feedback. Conversely, Type II feedback is applied to decrease the number of clauses that evaluate to 1 when the predicted output is higher than the target output ($y_o > \hat{y}_o$).

To stabilize learning, we use an activation probability function that makes the probability of giving a clause feedback proportional to the difference  between the predicted and target output (the error). That is, in the RTM, feedback to clauses is determined stochastically using the following activation probability function, $P_{act}$:

\begin{equation}\label{eq5}
P_{act} = \frac{K \times \mid y_o - \hat{y}_o \mid}{\hat{y}_{\mathrm{max}}} \;. 
\vspace{2mm}
\end{equation}
As seen, the magnitude of the function is adjusted with the constant \textit{K}. The resulting activation function reduces the oscillation of the the predicted value during the training process, stabilizing it around the target value.

The behavior of the RTM is studied in the following sections, in comparison with the CTM and MTM.

\section{Empirical Results}

\subsection{Experiment Setup}

We study the behavior of the RTM using six different datasets. These datasets have been constructed to facilitate empirical analysis of the optimality of RTM learning, with the underlying input-output mapping being known. Dataset I contains 2-bit feature input. The output is 100 times larger than the decimal value of the binary input (e.g., when the input is [1, 0], the output is 200). The training set consists of 8000 samples while the testing set consists of 2000 samples, both without noise. Dataset II contains the same data as Dataset I, except that the output of the training data is perturbed to introduce noise. For Dataset III we introduce 3-bit input, without noise, and for Dataset IV we have 3-bit input with noisy output. Finally, Dataset V has 4-bit input without noise, and Dataset VI has 4-bit input with noisy.

Each input feature have been generated independently with equal probability of $0$ and $1$ values, leading to a more or less uniform distribution of bit values.

In order to increase our understanding of the RTM, we investigate the effect the hyper-parameters $T$ and $s$  have on learning.

\vspace{3mm}
\textit{Experiment I}: We first study the effect varying \textit{T} has on performance for the different datasets.

\vspace{3mm}
\textit{Experiment II}: The effect of different $s$ values (controlling the number of sub-patterns) is further investigated for all of the datasets.

\vspace{3mm}
\textit{Experiment III}: We finally compare the RTM results with what can be achieved with  CTM and MTM.

\subsection{Results and Discussion}

\begin{figure*}[t]
\vspace{-2mm}
\centering
\includegraphics[width=12.3cm]{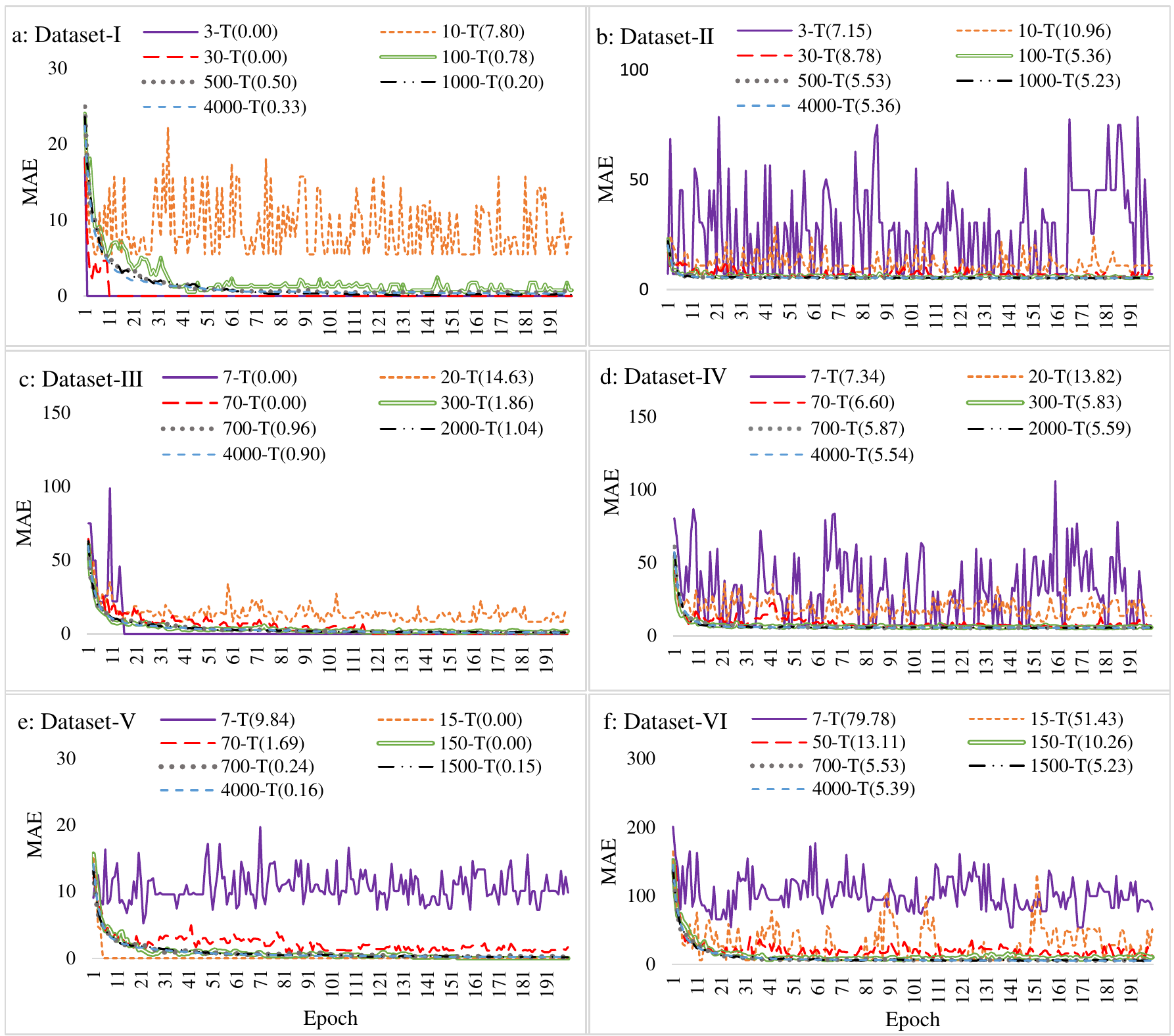}
\vspace{-7mm}
\caption{Training error over training epochs. Each dataset is processed with different \textit{T}.} \label{fig1}
\vspace{-3mm}
\end{figure*}

We use Mean Absolute Error (\textit{MAE}) to measure performance. Fig. \ref{fig1} plots error across 200 epochs, with learning influenced by different $T$ values. Fig. 1(a) shows the results for Dataset I, Fig. 1(b) reports results for Dataset II, and so on. MAE after 200 epochs is also given in brackets for each threshold in the legend.

From Fig. 1, we can observe that just 3 clauses (\textit{T} = 3) are enough to reduce error to zero for Dataset I, which can be explained by the noise-free data. Because the output value is decided by the number of clauses that output $1$, we require \emph{two} clauses with $\mathrm{TA}_1^1 = \{in\}$, $\mathrm{TA}_1^2 = \{ex\}$, $\mathrm{TA}_2^1 = \{ex\}$, and $\mathrm{TA}_2^2 = \{ex\}$ to capture the pattern (1 \ding{83}); see Phase 4 in Table~\ref{tab1}. Further, we need \emph{one} clause with $\mathrm{TA}_1^1 = \{ex\}$, $\mathrm{TA}_1^2 = \{ex\}$, $\mathrm{TA}_2^1 = \{in\}$, and $\mathrm{TA}_2^2 = \{ex\}$ to capture the pattern (\ding{83} 1). Here, \ding{83} means an input feature that can take an arbitrary value, either 0 or 1. These three clauses can collectively form any outputs for the Dataset I as shown in Table~\ref{clauses}. For instance, input (0 1) only activates the clause with $\mathrm{TA}_1^1 = \{ex\}$, $\mathrm{TA}_1^2 = \{ex\}$, $\mathrm{TA}_2^1 = \{in\}$, and $\mathrm{TA}_2^2 = \{ex\}$, which represents the pattern (\ding{83}~1). Accordingly, the RTM correctly computes the output, 100. Likewise, input (1 0) only activates the \emph{two} clauses with $\mathrm{TA}_1^1 = \{in\}$, $\mathrm{TA}_1^2 = \{ex\}$, $\mathrm{TA}_2^1 = \{ex\}$, and $\mathrm{TA}_2^2 = \{ex\}$, which represent the pattern (1  \ding{83}). Thus, the output 200 is correctly computed. All the clauses are activated when the input is (1 1) and therefore the output 300 is computed correctly as well.

We observe similar behaviour for Dataset III and Dataset V. More specifically, Dataset III requires \emph{seven} clauses to represent the three different patterns it contains, namely, (4 × (1 \ding{83} \ding{83}), 2 × (\ding{83} 1 \ding{83}), 1 × (\ding{83} \ding{83} 1)) \footnote{In this expression, \enquote{\emph{four} clauses to represent the pattern (1 \ding{83} \ding{83})} is written as \enquote{4 × (1 \ding{83} \ding{83})}}. Further, Dataset V requires \emph{fifteen} clauses to represent four different patterns it contains (8 × (1 \ding{83} \ding{83} \ding{83}), 4 × (\ding{83} 1 \ding{83} \ding{83}), 2 × (\ding{83} \ding{83} 1 \ding{83}), 1 × (\ding{83} \ding{83} \ding{83} 1)). As we can see from these 3 datasets, RTM can reach 0.00 for the training \textit{MAE} when \textit{T} is a multiplier of the minimum required clauses. For example, Dataset I can also be perfectly learned when there are 30 clauses.

\begin{table}[t]
\begin{center}
\centering
\caption{Computing output for different datasets by activating different clauses.}\label{clauses}
\vspace{-4mm}
\begin{tabular}{|c|c|c|}
\hline
\multicolumn{1}{|l|}{Dataset} & \multicolumn{1}{l|}{Output} & Required number of clauses to represent different patterns$^{\dag \dag}$ \\ \hline
\multirow{4}{*}{I}            & 0                           & \multicolumn{1}{r|}{None}                                         \\ \cline{2-3} 
                              & 100                         & \multicolumn{1}{r|}{1 × (\ding{83} 1)}    \\ \cline{2-3} 
                              & 200                         & \multicolumn{1}{r|}{2 × (1 \ding{83})}                                    \\ \cline{2-3} 
                              & 300                         & \multicolumn{1}{r|}{2 × (1 \ding{83}) + 1 × (\ding{83} 1) }                      \\ \hline
\multirow{8}{*}{III}            & 0                           & \multicolumn{1}{r|}{None }                                        \\ \cline{2-3} 
                              & 100                         & \multicolumn{1}{r|}{1 × (\ding{83} \ding{83} 1) }                                 \\ \cline{2-3} 
                              & 200                         & \multicolumn{1}{r|}{2 × (\ding{83} 1 \ding{83}) }                                   \\ \cline{2-3} 
                              & 300                         & \multicolumn{1}{r|}{2 × (\ding{83} 1 \ding{83}) + 1 × (\ding{83} \ding{83} 1)}        \\ \cline{2-3} 
                              & 400                         & \multicolumn{1}{r|}{4 × (1 \ding{83} \ding{83})}                            \\ \cline{2-3} 
                              & 500                         & \multicolumn{1}{r|}{4 × (1 \ding{83} \ding{83}) + 1 × (\ding{83} \ding{83} 1)}  \\ \cline{2-3} 
                              & 600                         & \multicolumn{1}{r|}{4 × (1 \ding{83} \ding{83}) + 2 × (\ding{83} 1 \ding{83})}  \\ \cline{2-3} 
                              & 700                         & \multicolumn{1}{r|}{4 × (1 \ding{83} \ding{83}) + 2 × (\ding{83} 1 \ding{83}) + 1 × (\ding{83} \ding{83} 1)} \\ \hline
\multirow{16}{*}{V}           & 0                           & \multicolumn{1}{r|}{None}                                         \\ \cline{2-3} 
                              & 100                         & \multicolumn{1}{r|}{1 × (\ding{83} \ding{83} \ding{83} 1)}  \\ \cline{2-3} 
                              & 200                         & \multicolumn{1}{r|}{2 × (\ding{83} \ding{83} 1 \ding{83})}   \\ \cline{2-3} 
                              & 300                         & \multicolumn{1}{r|}{2 × (\ding{83} \ding{83} 1 \ding{83}) + 1 × (\ding{83} \ding{83} \ding{83} 1)} \\ \cline{2-3} 
                              & 400                         & \multicolumn{1}{r|}{4 × (\ding{83} 1 \ding{83} \ding{83})}             \\ \cline{2-3} 
                              & 500                         & \multicolumn{1}{r|}{4 × (\ding{83} 1 \ding{83} \ding{83}) + 1 × (\ding{83} \ding{83} \ding{83} 1)}  \\ \cline{2-3} 
                              & 600                         & \multicolumn{1}{r|}{4 × (\ding{83} 1 \ding{83} \ding{83}) + 2 × (\ding{83} \ding{83} 1 \ding{83})}    \\ \cline{2-3} 
                              & 700                         & \multicolumn{1}{r|}{4 × (\ding{83} 1 \ding{83} \ding{83}) + 2 × (\ding{83} \ding{83} 1 \ding{83}) + 1 × (\ding{83} \ding{83} \ding{83} 1)} \\ \cline{2-3} 
                              & 800                         & \multicolumn{1}{r|}{8 × (1 \ding{83} \ding{83} \ding{83})} \\ \cline{2-3} 
                              & 900                         & \multicolumn{1}{r|}{8 × (1 \ding{83} \ding{83} \ding{83}) + 1 × (\ding{83} \ding{83} \ding{83} 1)}  \\ \cline{2-3} 
                              & 1000                        & \multicolumn{1}{r|}{8 × (1 \ding{83} \ding{83} \ding{83}) + 2 × (\ding{83} \ding{83} 1 \ding{83})} \\ \cline{2-3} 
                              & 1100                        & \multicolumn{1}{r|}{8 × (1 \ding{83} \ding{83} \ding{83}) + 2 × (\ding{83} \ding{83} 1 \ding{83}) + 1 × (\ding{83} \ding{83} \ding{83} 1)} \\ \cline{2-3} 
                              & 1200                        & \multicolumn{1}{r|}{8 × (1 \ding{83} \ding{83} \ding{83}) + 4 × (\ding{83} 1 \ding{83} \ding{83})} \\ \cline{2-3} 
                              & 1300                        & \multicolumn{1}{r|}{8 × (1 \ding{83} \ding{83} \ding{83}) + 4 × (\ding{83} 1 \ding{83} \ding{83}) + 1 × (\ding{83} \ding{83} \ding{83} 1)} \\ \cline{2-3} 
                              & 1400                        & \multicolumn{1}{r|}{8 × (1 \ding{83} \ding{83} \ding{83}) + 4 × (\ding{83} 1 \ding{83} \ding{83}) + 2 × (\ding{83} \ding{83} 1 \ding{83})}  \\ \cline{2-3} 
                              & 1500                        & \multicolumn{1}{r|}{8 × (1 \ding{83} \ding{83} \ding{83}) + 4 × (\ding{83} 1 \ding{83} \ding{83}) + 2 × (\ding{83} \ding{83} 1 \ding{83}) + 1 × (\ding{83} \ding{83} \ding{83} 1)}  \\ \hline
\end{tabular}
\vspace{-10mm}
\dag \dag \hspace{1mm}for example, \enquote{\emph{two} clauses to represent the pattern (1 \ding{83})} is written as \enquote{2 × (1 \ding{83})}
\end{center}
\vspace{-1mm}
\end{table}

However, when \textit{T} is not a multiplier of the minimum required clauses, RTM cannot align its output $y_o$ to the target output $\hat{y}_o$ during the training phase. For instance, by assigning \emph{four} clauses for Dataset I, the training will end up with e.g. allocating \emph{three} clauses to represent the pattern (1 \ding{83}) or \emph{two} clauses to represent the pattern (\ding{83} 1). As a result, one or more output values cannot be computed correctly. For example, if there are \emph{three} clauses for the pattern (1 \ding{83}) and \emph{one} clause for the pattern (\ding{83} 1) after training, input (1 0) activates the clauses that represent the pattern (1 \ding{83}), producing an incorrect output that is 300. Likewise, input (1 1) activates all four clauses to incorrectly compute the output 400.

As a strategy for problems where the number of clauses is unknown, and for real-world applications where noise plays a significant role, the RTM can be initialized with a much larger \textit{T}. Then, since the output, $y_o$, is a fraction of the threshold, \textit{T}, the error decreases. This behaviour is verified empirically in Fig.~\ref{fig1}, showing how increasing $T$ leads to reduced error.

\begin{figure}[t]
\vspace{-4mm}
\centering
\includegraphics[width=8.4cm]{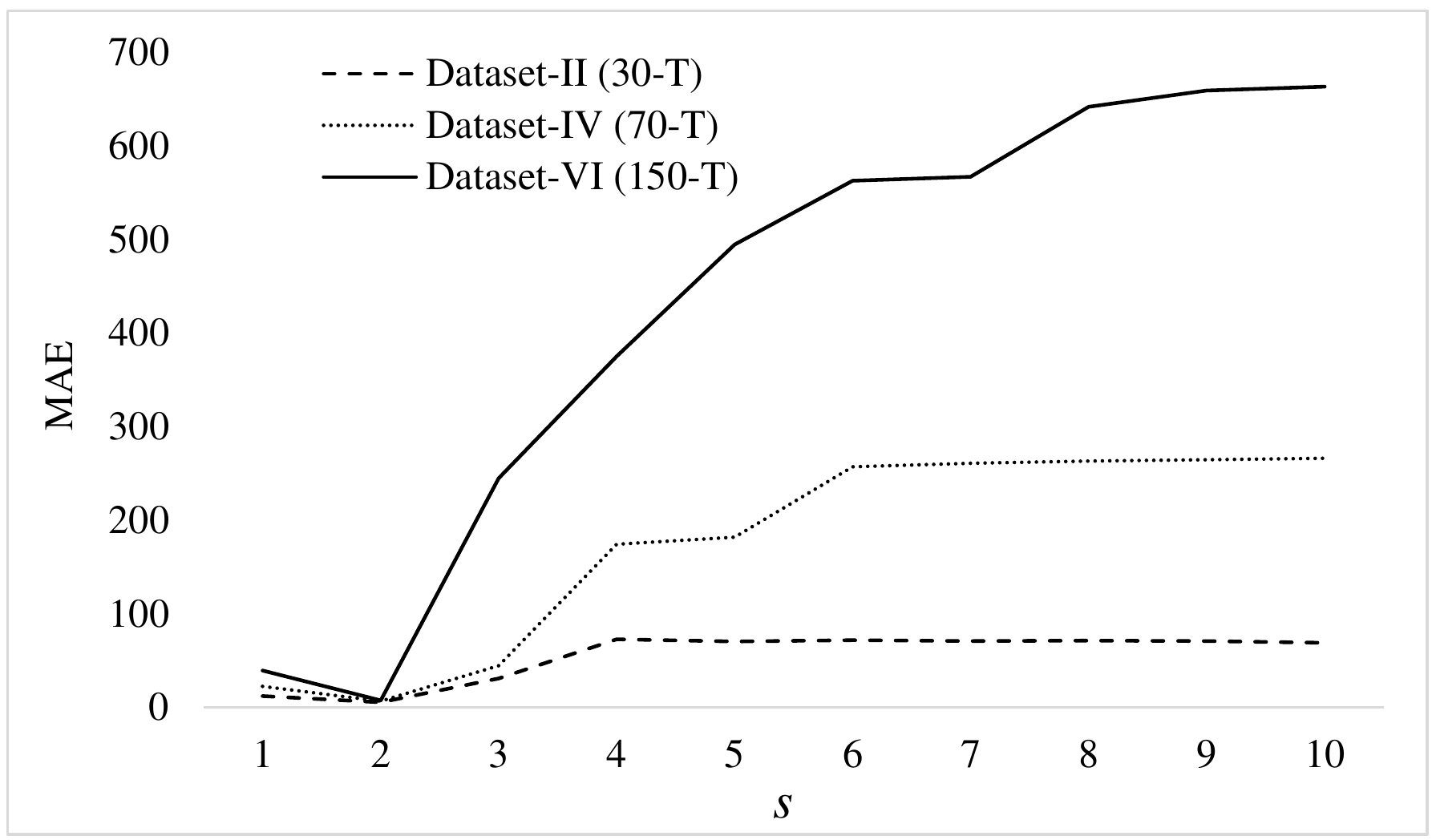}
\vspace{-3mm}
\caption{Variation of \textit{MAE} over different $s$ for fixed \textit{T}.} \label{fig2}
\vspace{-2mm}
\end{figure}

The effect of $s$ is studied by increasing it from $1.0$ to $10.0$ for Dataset II, Dataset IV, and Dataset VI, with fixed \textit{T}. Fig. \ref{fig2} shows the variation of \textit{MAE} over various $s$ values for noisy data. The \textit{MAE} decreases when $s$ increases from $1.0$ to $2.0$. After $2.0$, \textit{MAE} increases, and then stabilizes after a while.

\begin{figure}[t]
\centering
\includegraphics[width=4.4cm]{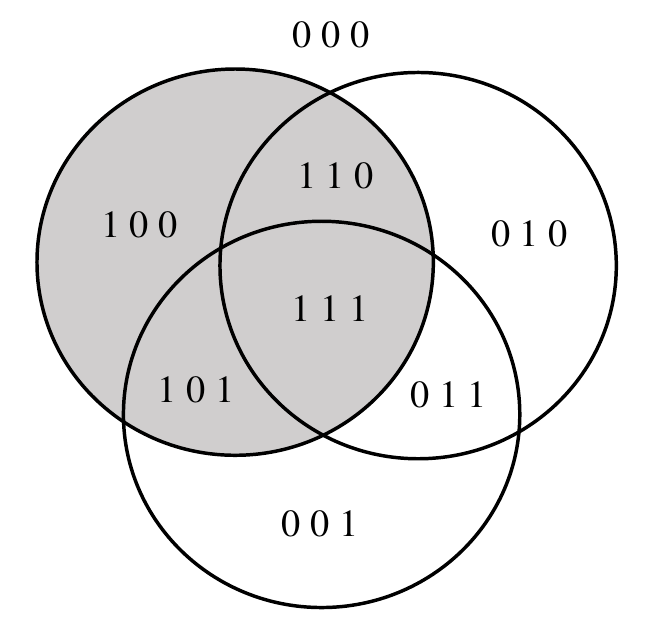}
\vspace{-3mm}
\caption{Pattern distribution for the 3-bits input datasets.} \label{fig3}
\vspace{-5mm}
\end{figure}

For all of the datasets considered here, the optimum $s$, where the RTM learns the datasets with minimum \textit{MAE}, is equal to $2.0$. The reason can be explained with the aid of Fig.~\ref{fig3}, where one sees the distribution of patterns when the dataset has 3 input bits. 

The occurrence probability of any of the 3-bit patterns is $\frac{1}{8}$ since there are overall 8 unique patterns. However, to capture the pattern (1 \ding{83} \ding{83}) (shaded area), according to the TM dynamics \cite{Ole1}, $\frac{1}{s}$ should be equal to the probability of the considered pattern, which is $\frac{4}{8} (=\frac{1}{2})$. Hence, $s$ should be $2$. For instance, if someone assigns $s = 4$, clauses will start to learn much finer patterns, such as (1 0 \ding{83}), (1 1 \ding{83}), and (0 1 \ding{83}). This significantly increases the number of clauses needed to capture the sub-patterns. This is also the case for Dataset II and Dataset VI. Then, the probability that (1 \ding{83}) occurs is $\frac{2}{4} (=\frac{1}{2})$ and the probability that (1 \ding{83} \ding{83} \ding{83}) occurs is $\frac{7}{14} (=\frac{1}{2})$.

To compare the performance of the RTM with CTM and MTM, each model is tested with different \textit{T} values. The training and testing \textit{MAE} for all the cases are summarized in Table \ref{tab3} and \ref{tab4}, respectively.

\begin{table*}[h]
\vspace{-5mm}
\begin{center}
\centering
\newcolumntype{P}[1]{>{\centering\arraybackslash}p{#1}}
\caption{Training \textit{MAE} after 200 training epochs with different \textit{T} on various methods.}\label{tab3}
\vspace{-3mm}
\renewcommand{\arraystretch}{1.2}
\begin{tabular}{P{4mm}P{4mm}c|P{7mm}|P{7mm}|P{7mm}|P{7mm}|P{7mm}|P{7mm}|P{7mm}|P{8mm}|P{8mm}|P{8mm}|P{8mm}|P{8mm}|}
\cline{4-15}
                                               &                        &     & \multicolumn{7}{c|}{\textbf{RTM}} & \multicolumn{2}{c|}{\textbf{CTM}} & \multicolumn{3}{c|}{\textbf{MTM}} \\ \cline{3-15} 
                                               & \multicolumn{1}{l|}{}  & \textit{T} & 3	& 10 & 30 &	100 & 500 &	1000 & 4000 & 6 & 8000 & 1000 &	10000 &	16000 \\ \hline  
\multicolumn{1}{|l|}{\multirow{8}{*}{\rotatebox[origin=c]{90}{\textbf{Dataset}}}} & \multicolumn{1}{l|}{\textbf{1}} & \textit{MAE} & 0.0 & 7.8 & 0.0 & 0.8 & 0.5 & 0.2 & 0.3 & 7.7 & 0.0 & 0.0 & 0.0 & 0.0 \\ \cline{2-15} 
\multicolumn{1}{|l|}{}                         & \multicolumn{1}{l|}{\textbf{2}} & \textit{MAE} & 7.2 & 11.0 & 8.8 & 5.4 & 5.5 & 5.2 & 5.4 & 11.1 & 24.1 & 8.4 & 7.1 & 7.9 \\ \cline{2-15}  
\multicolumn{1}{|l|}{}                         & \multicolumn{1}{l|}{}  & \textit{T} & 7 & 20 & 70 &	300 & 700 &	2000 & 5000 & 14 & 8000 & 2000 & 10000 & 16000 \\ \cline{2-15}  
\multicolumn{1}{|l|}{}                         & \multicolumn{1}{l|}{\textbf{3}} & \textit{MAE} & 0.0 & 14.6 & 0.0 & 1.9 & 1.00 &	1.0 & 0.9 &	0.0 & 0.0 &	18.3 & 0.0 & 0.0 \\ \cline{2-15}  
\multicolumn{1}{|l|}{}                         & \multicolumn{1}{l|}{\textbf{4}} & \textit{MAE} & 7.4 & 13.8 & 6.6 & 5.8 & 5.9 & 5.6 & 5.5 & 111.3 & 13.3 & 14.2 & 8.8 & 8.4 \\ \cline{2-15} 
\multicolumn{1}{|l|}{}                         & \multicolumn{1}{l|}{}  & \textit{T} & 7 & 15 & 70 &	150 & 700 &	1500 & 4000 & 30 & 8000 & 4000 & 10000 & 16000 \\ \cline{2-15} 
\multicolumn{1}{|l|}{}                         & \multicolumn{1}{l|}{\textbf{5}} & \textit{MAE} & 9.8 & 0.0 &	1.7 & 0.0 &	0.2 & 0.2 &	0.2 & 149.7 & 158.7 & 373.1 & 0.0 &	0.0 \\ \cline{2-15} 
\multicolumn{1}{|l|}{}                         & \multicolumn{1}{l|}{\textbf{6}} & \textit{MAE} & 79.8 & 51.4 & 13.1 & 10.3 &	5.5 & 5.3 &	5.4 & 181.5 & 96.4 & 449.9 & 8.0 & 7.8 \\ \hline 
\end{tabular}
\vspace{3mm}
\end{center}
\end{table*}

\begin{table*}[h]
\vspace{-8mm}
\begin{center}
\centering
\newcolumntype{P}[1]{>{\centering\arraybackslash}p{#1}}
\caption{Testing \textit{MAE} for different \textit{T} on various methods.}\label{tab4}
\vspace{-3mm}
\renewcommand{\arraystretch}{1.2}
\begin{tabular}{P{4mm}P{4mm}c|P{7mm}|P{7mm}|P{7mm}|P{7mm}|P{7mm}|P{7mm}|P{7mm}|P{8mm}|P{8mm}|P{8mm}|P{8mm}|P{8mm}|}
\cline{4-15}
                                               &                        &     & \multicolumn{7}{c|}{\textbf{RTM}} & \multicolumn{2}{c|}{\textbf{CTM}} & \multicolumn{3}{c|}{\textbf{MTM}} \\ \cline{3-15} 
                                               & \multicolumn{1}{l|}{}  & \textit{T} & 3	& 10 & 30 &	100 & 500 &	1000 & 4000 & 6 & 8000 & 1000 &	10000 &	16000 \\ \hline 
\multicolumn{1}{|l|}{\multirow{8}{*}{\rotatebox[origin=c]{90}{\textbf{Dataset}}}} & \multicolumn{1}{l|}{\textbf{1}} & \textit{MAE} & 0.0 & 7.6 &	0.0 & 0.8 &	0.5 & 0.2 &	0.3 & 9.0 &	0.0 & 0.0 &	0.0 & 0.0 \\ \cline{2-15} 
\multicolumn{1}{|l|}{}                         & \multicolumn{1}{l|}{\textbf{2}} & \textit{MAE} & 5.0 & 10.6 & 7.1 &	1.2 & 2.7 &	1.6 & 1.8 &	9.4 & 25.3 & 7.5 & 5.4 & 7.0 \\ \cline{2-15} 
\multicolumn{1}{|l|}{}                         & \multicolumn{1}{l|}{}  & \textit{T} & 7 & 20 & 70 &	300 & 700 &	2000 & 5000 & 14 & 8000 & 2000 & 10000 & 16000 \\ \cline{2-15} 
\multicolumn{1}{|l|}{}                         & \multicolumn{1}{l|}{\textbf{3}} & \textit{MAE} & 0.0 & 14.2 & 0.0 &	2.1 & 1.0 &	1.2 & 1.0 &	0.0 & 0.0 &	22.0 & 0.0 & 0.0 \\ \cline{2-15} 
\multicolumn{1}{|l|}{}                         & \multicolumn{1}{l|}{\textbf{4}} & \textit{MAE} & 5.0 & 14.5 & 4.2 &	3.3 & 3.4 &	1.9 & 2.7 &	98.5 & 12.5 & 16.0 & 8.7 & 8.3 \\ \cline{2-15} 
\multicolumn{1}{|l|}{}                         & \multicolumn{1}{l|}{}  & \textit{T} & 7 & 15 & 70 &	150 & 700 &	1500 & 4000 & 30 & 8000 & 4000 & 10000 & 16000 \\ \cline{2-15} 
\multicolumn{1}{|l|}{}                         & \multicolumn{1}{l|}{\textbf{5}} & \textit{MAE} & 9.9 &	0.0 & 1.8 &	0.0 & 0.3 &	0.2 & 0.2 &	154.6 &	155.5 &	372.9 &	0.0 & 0.0 \\ \cline{2-15} 
\multicolumn{1}{|l|}{}                         & \multicolumn{1}{l|}{\textbf{6}} & \textit{MAE} & 78.0 &	50.1 & 12.5 & 8.5 &	3.5 & 2.7 &	2.8 & 191.3 & 102.4 & 431.3 & 6.9 &	6.7 \\ \hline 
\end{tabular}
\end{center}
\vspace{-8mm}
\end{table*}

The training and testing \textit{MAE} reach zero when the RTM operates with noise free data and when \textit{T} equals the optimum clauses required. When the optimum \textit{T} is unknown, and when data is noisy, applying a higher \textit{T} is beneficial. As an example, Dataset III, which has 3 bits as inputs, can be perfectly learned with \textit{T} equal to 7 and 70. For the same dataset, RTM acquires training \textit{MAE} $1.0$ with \textit{T} equaling 700, which is better than the \textit{MAE} of $14.2$ obtained when \textit{T} equals $20$.

For CTM, the outputs are converted to bits and each bit position is then trained and predicted separately. According to the training and testing \textit{MAE} in Table 4 and 5, CTM works better with less complex datasets such as Dataset I and Dataset III. However, with a higher number of inputs and with noisy training data, performance decreases. 

MTM requires a large number of clauses by nature when it works with continuous outputs since it has to consider all possible values from 0 to $\hat{y}_\mathrm{max}$ as distinct classes (e.g. 300 classes for Dataset I and Dataset II, and 700 classes for Dataset III and Dataset IV). According to the training and testing \textit{MAE} in the Tables 4 and 5, MTM requires roughly 3 clauses or more per class. For instance, the features in Dataset I can be learned with 1000 clauses, yet that amount is insufficient for Dataset III and Dataset V. Note that the noise free datasets can be learned perfectly with 10000 or more clauses. However, this accuracy gain is accompanied with a larger computational cost. 

Overall, RTM obtains the best training and testing \textit{MAE} for both noisy and noise free data with a smaller number of clauses compared with the CTM and MTM.  Dataset II, Dataset IV, and Dataset VI are more similar to real-world datasets by being noisy. The minimum \textit{MAE} values obtained by RTM for these three Datasets are 1.6, 1.9, and 2.7, respectively. The average of these minimum \textit{MAE} values (2.07) is approximately 20 and 3.5 times lower than the averages obtained with CTM and MTM, respectively. In terms of the number of clauses required to achieve the above results, RTM utilizes 1000 clauses, while CTM and MTM utilize 8 and 16 times more clauses than that. This difference is characteristic for RTM -- it provides better \textit{MAE} with less computational power. 

\section{Conclusion}

In this paper we proposed the Regression Tsetlin Machine (RTM), a novel variant of the Classic Tsetlin Machine that supports continuous output in regression problems. In RTM, the polarities in clauses were removed and the total clause output was normalized to produce continuous output predictions. The number of clauses to receive the feedback in RTM was decided stochastically using a linear activation probability function. The prediction power of this novel approach was studied using six different datasets, with noise free and noisy training data. Our empirical results showed significantly better performance of RTM compared with CTM and MTM, both in terms of training and the testing error, as well as the computational power required. 

Potential applications for RTM can be weather prediction, sales forecasting, stock predictions, energy forecasting, and outbreak forecasting, to name a few. In our future work, we will evaluate RTM on the aforementioned applications and performance will be compared with conventional machine learning methods.

\end{document}